\documentclass[10pt,twocolumn,letterpaper]{article}
\usepackage{wacv}
\usepackage{times}
\usepackage{tablestyles}
\usepackage{tabularx}
\usepackage{epsfig}
\usepackage{graphicx}
\usepackage{amsmath}
\usepackage{amssymb}
\usepackage{subcaption}
\usepackage{rotating}
\usepackage{pdflscape}
\usepackage{cite}
\usepackage{graphicx}
\usepackage{float}
\usepackage{pifont}
\usepackage{amsmath}
\usepackage{mathtools}
\graphicspath{ {./} }
\usepackage{comment}
\usepackage[normalem]{ulem}
\useunder{\uline}{\ul}{}
\usepackage[tableposition=top]{caption}
\usepackage{textcomp}
\usepackage{authblk}
\renewcommand\vec{\mathbf}

\wacvfinalcopy 
\ifwacvfinal
\def\assignedStartPage{1}
\fi

\ifwacvfinal
\usepackage[pagebackref=true,breaklinks=true,colorlinks,bookmarks=false]{hyperref}
\else
\usepackage[pagebackref=true,breaklinks=true,colorlinks,bookmarks=false]{hyperref}
\fi

\ifwacvfinal
\else
\fi

\begin{document}

\title{O\textsubscript{2}A: One-shot Observational learning with Action vectors}
\author
{Leo Pauly, Wisdom C. Agboh, David C. Hogg, Raul Fuentes \\
University of Leeds, United Kingdom \\
{\tt\small}
}
\maketitle


\begin{abstract}

We present O\textsubscript{2}A, a novel method for learning to perform robotic manipulation tasks from a single (one-shot) third-person demonstration video. To our knowledge, it is the first time this has been done for a single demonstration. The key novelty lies in pre-training a feature extractor for creating a perceptual representation for actions that we call \lq \textit{action vectors}\rq.  The action vectors are extracted using a 3D-CNN model pre-trained as an action classifier on a generic action dataset.  The distance between the action vectors from the observed third-person demonstration and trial robot executions is used as a reward for reinforcement learning of the demonstrated task. We report on experiments in simulation and on a real robot, with changes in viewpoint of observation, properties of the objects involved, scene background and morphology of the manipulator between the demonstration and the learning domains. O\textsubscript{2}A outperforms baseline approaches under different domain shifts and has comparable performance with an oracle (that uses an ideal reward function). Videos of the results, including demonstrations, can be found in our: \href{https://leopauly.github.io/s2l/my_docs/stage1}{project-website. \footnote{We have anonymized the project website by removing all the identification details for the review purpose. It currently contains only the videos of the results. Full link: \url{https://leopauly.github.io/s2l/my_docs/stage1}}}

\end{abstract}


\section{Introduction}

\begin{figure}[t]
   \centering
   \includegraphics[scale=.27]{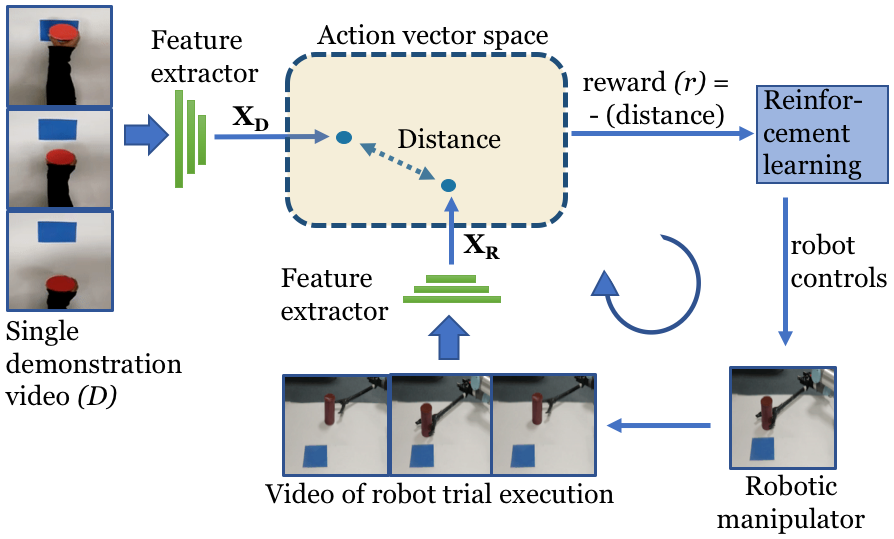}
    \caption{Overview of O\textsubscript{2}A method. A 3D-CNN action vector extractor is used to extract action vectors $\vec{X_{D}}$ and $\vec{X_{R}}$ from the video clips of the demonstration and robot trial execution respectively. A reward function is used to compare  $\vec{X_{D}}$ and $\vec{X_{R}}$ in the action vector space, generating a reward signal ($r$) based on their closeness. The reinforcement learning algorithm then iteratively learns an optimal control policy by maximizing this reward signal, thus enabling observational learning.}
    \label{fig:Overview_of_the_proposed_system}
\end{figure}

Learning new manipulation tasks has always been challenging for robotic systems, whether it is a simple mobile manipulator or a complex humanoid robot. Programming manually step by step \cite{finkel1975overview} is one of the earlier solutions to this problem. But this is labour intensive, requires specialist expertise and lacks autonomy. It is therefore not suitable for consumer robots and fully autonomous systems. Learning from Demonstrations (LfD)~\cite{atkeson1997robot} is a potential solution to this problem. It requires only demonstrations of the task for the robot to learn from. Even though LfD has been studied widely, most previous works have stayed within the~\lq Imitation Learning\rq\cite{duan2017one, pathak2018zero, peng2018deepmimic,argall2009survey} paradigm, where demonstrations are made from an egocentric perspective, either visually or kinesthetically. This requires the inconvenience of kinaesthetic guidance or teleoperation and also the rich source of third-person demonstrations available on the internet cannot be used. Therefore, in this paper we study the problem of LfD under the ~\lq Observational Learning\rq ~\cite{Torabi2019,pauly2018defining,Borsa2019,psychology} paradigm, where the demonstrations are observed as a third-person. This introduces the key challenge in observational learning, the shift between the demonstration and the learning domains. The domain shift can arise due to changes in viewpoints of observation, properties of objects used, scene background or morphology of the manipulator performing the task. 

In this paper we present O\textsubscript{2}A (\textbf{O}ne-shot \textbf{O}bservational learning with \textbf{A}ction vectors), for one-shot observational learning of robotic manipulation tasks under different domain shifts. One-shot learning here means that only a single demonstration of the new task is required for learning. (Note that, it does not refer to the number of trial and error executions by the robot during learning from that single demonstration). We use an abstract perceptual representation: the \lq action vector\rq~, which is the task-aware and domain-invariant representation of the action in a video. The action vector is extracted using a 3D-CNN \cite{tran2015learning}, pre-trained on a generic action dataset as an action classifier (we use UCF101 \cite{soomro2012ucf101} as the pre-training dataset for our experiments). Through our evaluation on a new \lq Leeds Manipulation Dataset\rq~(LMD), we show that the pre-trained action vector extractor can generalise to unseen manipulation tasks. The action vectors from the demonstration and trial robot execution video clips are then compared to generate a reward for the reinforcement learning algorithm. The algorithm learns an optimal control policy that performs the demonstrated task. Our experiments in simulation (with reaching, pushing tasks) and on a real robot (with pushing, hammering, sweeping, striking tasks) show that O\textsubscript{2}A can perform well under different domain shifts. Our contributions can be summarised as follows:

\begin{itemize}
    \item Implementing for the first time, a method for observational learning of robotic manipulation tasks from a single demonstration.
    \item O\textsubscript{2}A can handle shifts between the demonstration and the learning domains, caused by changes in viewpoint of observation, object properties, morphology of the manipulator and scene background.
    \item And finally,  we pre-train the action vector extractor on a generic action dataset instead of task-specific manipulation videos. The extractor generalises to unseen manipulation tasks by learning the shared underlying visual dynamics. 
\end{itemize}

Upcoming sections are arranged as follows: Section~\ref{literature} discusses related works, Section~\ref{proposed_method} formulates the problem and describes the proposed method, Sections~\ref{sec:exp_results_action} and \ref{sec:exp_results_sim_robot} report on experiments conducted and finally Section~\ref{conclusion} presents the conclusions.

\section{Related Work} 
\label{literature}

\textbf{Observational learning:}
Origins of observational learning of robotic manipulation tasks can be traced back to works from the 1990s \cite{Kuniyoshi1994,See1993,Suehiro1994a}. Most of the early methods required assistance in observing the demonstrations. This assistance was provided by motion capture systems\cite{Field2009,Ijspeert2001,Ijspeert2002}, visual detectors\cite{Ramirez-Amaro2017a,Sieb2019,Zhang2019}, trackers\cite{Dragan2012,Gupta2016,Dillmann2004} or a combination of the above \cite{Kuniyoshi1994}. However, the entities to be tracked or detected must be known beforehand and only demonstrations using these entities can be learned. 

With the advent of deep learning \cite{goodfellow2016deep,lecun2015deep}, it was possible to learn visual features characterising the task directly from raw RGB videos. The features are extracted from raw videos using a variety of methods: deep metric learning \cite{sermanet2018time}, generative adversarial learning \cite{stadie2017}, domain translation \cite{Liu2018,Smith2019,Sharma2019}, transfer learning \cite{Sharma2018,sermanet2016unsupervised}, action primitives \cite{Jia2020}, predictive modelling \cite{Tow2017}, video to text translation \cite{Yang2019} and meta-learning \cite{yu2018one,yu2018two}. A detailed comparison of these methods is given in the supplementary material Section A.

These methods have two main limitations: (1) Requirement of a large number of demonstrations for learning new tasks: The feature extractors are trained separately for each of the new task to be learned. Hence demonstration videos are to be collected in substantial numbers for each task. In contrast, our method requires only a single demonstration (hence one-shot) to learn a new task, since pre-trained feature extractors are used. (2) Constrained domain shifts: In existing approaches, assumptions are made regarding shift between learning and demonstration domains. For example viewpoint of observation is fixed \cite{sermanet2016unsupervised} or manipulators with similar morphologies \cite{Liu2018} are used. Our method O\textsubscript{2}A, does not make any such assumptions and can learn under unconstrained domain shifts.

\textbf{Pre-training with large generic datasets:}  Pre-training on large generic datasets has become common in the fields of computer vision and natural language processing.  Models are first pre-trained on a large generic dataset(s) in a supervised or unsupervised manner. After pre-training, the models are used to solve downstream tasks with minimum/no fine-tuning. Generic language models such as ELMo \cite{peters2018deep}, GPT \cite{radford2018improving,radford2019language,brown2020language}, BERT \cite{devlin2018bert} have shown success in solving several downstream language processing tasks. Similarly, ImageNet models \cite{xie2018pre}, Image-GPT \cite{chen2020generative}, BiT models \cite{kolesnikov2019big} have demonstrated that this approach can be applied for computer vision problems as well. We introduce a similar  concept into visual robotic manipulation.  Action vector extractors are pre-trained using a large generic action dataset and then generalised to manipulation tasks for observational learning.

\textbf{Mirror Neurons:} Neuroscience studies \cite{rizzolatti2005mirror,rizzolatti2004mirror,cattaneo2009mirror,lago2014role} show the presence of  \lq mirror neurons\rq~in humans, that produce task-aware and domain-invariant representations of the actions observed. They are used for both action recognition (perception) and observational learning (action execution). Inspired by this dual role of mirror neurons, we pre-train the action vector extractor as an action classifier. This pre-training for action classification (recognition), will teach the model to extract task-aware and domain-invariant representations of actions from input videos.

\section{Proposed method} \label{proposed_method}

\subsection{Action vectors}
\label{sec:action_vectors}

Action vectors are the core of the O\textsubscript{2}A method. An action vector is the abstract task-aware and domain-invariant perceptual representation of the action being carried out in a video. In O\textsubscript{2}A, the action vector extraction is based on the following two assumptions:

(1) The spatio-temporal features generated by the final layers of an action classifier pre-trained on a generic action dataset, are domain-invariant and task-aware. The features from the videos depicting similar actions should be identical irrespective of the domain in which they are recorded. The assumption is reasonable since the action classifier makes use of the same layer outputs to identify actions, independently of different camera angles, varying scene backgrounds, illumination conditions, actors / manipulators, object appearances, interactions,  pose and scale. 

(2) The action vector extraction model pre-trained on a generic dataset can generalise to unseen manipulation tasks used in robotic observational learning. The intuition is that the underlying visual dynamics between generic action datasets and manipulation tasks are the same. For example, it is the same physical laws of dynamics governing object interactions, both for a cricket shot as well as a robot striking cubes.

Section \ref{sec:exp_results_action} shows results that validate these critical assumptions.

\subsubsection{Network architecture and dataset}

Our 3D-CNN model consists of eight 3D convolutional layers, five 3D maxpooling layers and three fully connected layers. The ReLU \cite{nair2010rectified} activation function is used for all the convolutional and fully connected layers except the final layer, where a linear activation function followed by a Softmax function is used. The layer wise network architecture along with the kernel sizes, input and output  dimensions are given in the supplementary material Section B. We use the UCF101 action dataset as the generic dataset for our experiments. It consists of 13320 real world action videos from YouTube each lasting around 7 seconds on average, classified into 101 action categories. The dataset has a large diversity both in terms of variety of actions and domain settings within the same class videos. 

\subsubsection{Pre-training action vector extractors}

For pre-training, we first uniformly downsample UCF101 videos in time into 16 frames for providing a fixed-length representation for each video clip. We also resize videos into 112 x 112 pixels to standardize the size. We apply the same pre-processing steps to videos of demonstrations and robot trial executions for action vector extraction during observational learning. These downsampled and resized videos are then used for training the model for action classification from scratch.  The training details are given in the supplementary material Section B. The trained model will be referred to as \lq NN:UCF101\rq~hereafter. 

After training, we use features from one of the final layers of NN:UCF101 as the action vector. Our experiment (reported in the supplementary material Section C) shows that the features from layers pool5 (size: 8192) and fc6 (size: 4096) are best suited to be used as the action vector. We report results, both when the features from pool5 and fc6 layers are used as the action vector in this paper.

\subsection{One-Shot observational learning}

The overview of O\textsubscript{2}A is shown in Fig.~\ref{fig:Overview_of_the_proposed_system}. The robot views both the demonstration and its own trial executions from a camera mounted in a fixed position above the manipulator. With reference to Fig~\ref{fig:Overview_of_the_proposed_system}, let $D$ be the single demonstration video clip of a task to be learned. We extract the $n$-dimensional action vectors $\vec{X_{D}}$ and $\vec{X_{R}}$ from the demonstration video $D$ and the video clip of a trial robot execution respectively. The reward ($r$) for the reinforcement learning is then calculated as the negative of the euclidean distance between action vectors  $\vec{X_{D}}$ and $\vec{X_{R}}$ as given below: 
\begin{align}
\label{reward_equation}
r = -||\vec{X_D}-\vec{X_R}||_2 
\end{align}
Thus the reward directly measures the closeness of the actions in the demonstration and of the robot trial execution. The reinforcement learning will then maximize this reward function to learn an optimal control policy. This optimal control policy will enable the robotic manipulator to perform the demonstrated task.

\subsubsection{Reinforcement learning of the task}
Any reinforcement learning algorithm can be used with our method. In the simulation experiment, we use the Deep Deterministic Policy Gradient (DDPG)~\cite{Lillicrap2015ContinuousCW} to estimate the optimal control policy. The states used by the control policy are instantaneous visual observations of the environment (as observed by the robotic system). We make use of a VGGNet pre-trained on ImageNet~\cite{simonyan2014very} for converting raw RGB images into visual state features. The 4608 long feature obtained from the last convolutional layer of the VGG-16 network is used as the instantaneous state representation.

Reinforcement learning in real robots is an active area of research and remains a challenging problem. So we use a manipulation planning algorithm, the Stochastic Trajectory Optimisation (STO)\cite{Agboh_trajopt,kalakrishnan2011stomp}, for the real robot experiment.  STO generates an optimal control sequence by iteratively improving on the previous sequence guided by our reward function. The cost function $C$, to be minimized is calculated as: $C = r^{2}$. 

\section{Action vector analysis}
\label{sec:exp_results_action}

\begin{figure}[!h]
    \centering
    \includegraphics[scale=.215]{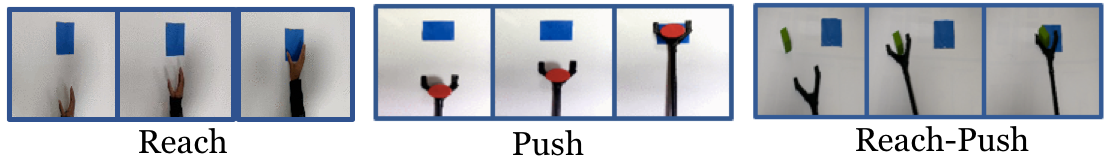}
    \caption{Snapshots of sample videos from LMD. Identical looking task classes are used to make the task differentiation more challenging.}
\label{fig:Leeds_dataset2}
\end{figure}

In this section, we aim to validate our assumptions for the proposed action vector extraction method explained in Section \ref{sec:action_vectors}. First we collect a manipulation tasks dataset, the \lq Leeds Manipulation Dataset (LMD)\rq. Note that this dataset is only used for evaluation and not used during training of the action vector extractor. LMD consists of videos of three different manipulation tasks: reach, push and reach-push, examples of which are shown in Figure \ref{fig:Leeds_dataset2}. The task videos are collected directly with a human hand and by using tools resembling robotic manipulators/end effectors. Each class consists of 17 videos with variations in viewpoint, object properties, scene background and morphology of manipulator within each class. Note that, identical looking task classes were carefully selected and same set of objects and manipulators were used across tasks for collecting videos. These choices are deliberate to make the task differentiation more challenging. Under these circumstances, only an efficient action vector extractor can produce task-aware and domain-invariant action vectors for different task classes in LMD.

\subsection{Class similarity scores}

Here we calculate the interclass and intraclass similarity scores for different classes of LMD in the action vector space. For that, we extract action vectors from pool5 and fc6 layers of the NN:UCF101 model, for all the 51 videos in LMD. The Baseline-R is obtained using  features from pool5 layer of the same NN:UCF101 model but initialised with random weights. The similarity score between a pair of action vectors, is shown as the cosine of the angle between them. The similarity scores are bounded by $[-1,1]$ with $-1$ indicating diametrically opposite vectors and $1$ indicating coinciding vectors.  

The results are tabulated in Table \ref{table:sim_scores}. For each chosen feature layer, the diagonal values represent the average of similarity scores between pairs of action vectors from the same class. And the non-diagonal values are the average of similarity scores between pairs of action vectors from different classes. The diagonal values are greater than the rest of the values indicating adequate task-awareness and domain-invariance for the action vectors extracted. The only exception is for layer fc6 where a greater inter-class similarity score is observed between reach and push classes than the intraclass similarity score for reach class. Provided that both tasks are extremely similar, these results are promising.

\begin{table}[!t]
\centering
\tablestyle
\rmfamily
\caption{Class similarity scores. The  intraclass similarity  (diagonal  values)  are  greater  than  the  rest  of the values, indicating adequate task-awareness and domain-invariance.}
\label{table:sim_scores}
\begin{tabular}{p{2cm}m{1.2cm}m{1.2cm}c}
\hline
\textbf{Baseline-R (Random weights)}              & \textbf{Reach}  & \textbf{Push}   & \textbf{Reach-Push} \\ \hline
\textbf{Reach}         & 0.9873 & 0.9870 & 0.9870        \\ 
\textbf{Push}          & 0.9870 & 0.9874 & 0.9868        \\ 
\textbf{Reach-Push} & 0.9870 & 0.9868 & 0.9889        \\ \hline
\end{tabular}
\newline

\centering
\tablestyle
\rmfamily
\begin{tabular}{p{2cm}m{1.2cm}m{1.2cm}c}
\hline
\textbf{NN:UCF101 (pool5)}              & \textbf{Reach}           & \textbf{Push}            & \textbf{Reach-Push}   \\ \hline
\textbf{Reach}         & \textbf{0.7391} & 0.7371          & 0.6897          \\ 
\textbf{Push}          & 0.7371          & \textbf{0.7547} & 0.6852          \\ 
\textbf{Reach-Push} & 0.6897          & 0.6852          & \textbf{0.7578} \\ \hline
\end{tabular}
\newline

\centering
\tablestyle
\rmfamily
\begin{tabular}{p{2cm}m{1.2cm}m{1.2cm}c}
\hline
\textbf{NN:UCF101 (fc6)}              & \textbf{Reach}                               & \textbf{Push}            & \textbf{Reach-Push}   \\ \hline
\textbf{Reach}         & { 0.4994} & 0.5001          & 0.4052          \\ 
\textbf{Push}          & \textbf{0.5001 }                             & \textbf{0.5352} & 0.4022          \\ 
\textbf{Reach-Push} & 0.4052                              & 0.4022          & \textbf{0.4978} \\ \hline
\end{tabular}
\newline

\end{table}

\begin{figure*}[!t]
    \centering
    \includegraphics[scale=.47]{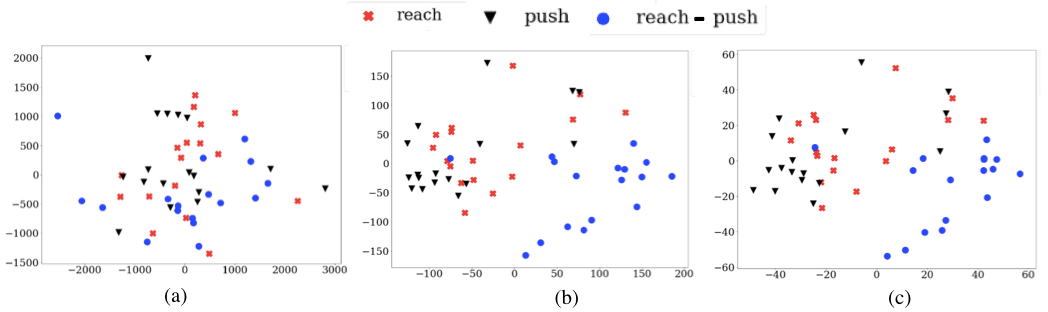}
    \caption{Visualising LMD for (a) Baseline-R and using action vectors from (b) pool5 and (c) fc6 layers  of NN:UCF101 model. We can see the clustering of the action vectors into different task classes.}
\label{fig:vis_all}
\end{figure*}

We also visualize these action vectors from LMD, projected into 2D using PCA, which are shown in Figure \ref{fig:vis_all}. The clustering of action vectors from the same classes, when compared to the Baseline-R is evident. This further indicates the domain-invariance and task-awareness of our action vectors. It must be noted that this visualisation collapses the vectors, of much greater dimensions, into a 2D space, which might be causing some of the \lq artificial\rq~ overlaps.

The class similarity scores and visualization shows that our pre-trained  action vector extractor can generalise to unseen manipulation tasks. In the next section we show how the action vector is used for observational learning and how well O\textsubscript{2}A performs under different domain shifts.

\section{Robotic experiments}
\label{sec:exp_results_sim_robot}

\begin{figure*}[!t]
   \centering
   \smallskip
   \medskip
   \includegraphics[trim={0cm 0cm 0cm 0cm}, scale=.36]{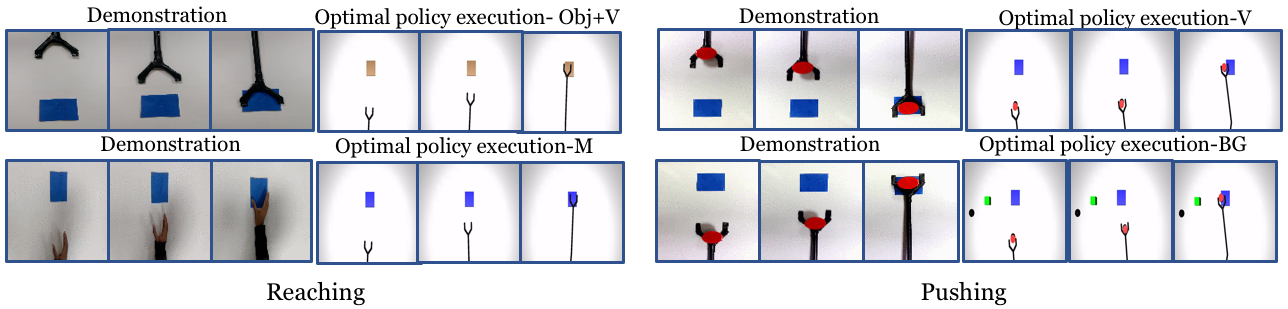}  
   \captionsetup{justification=centering}
    \caption{Snapshots of the demonstration and the execution of corresponding learned policies in the simulation experiment for selected domain shifts. (Results shown for action vectors extracted from pool5 layer of NN:UCF101 model).}
    \label{fig:demons_n_learned}
\end{figure*}

\begin{table}[!t]
\tablestyle[default]
\rmfamily
\centering
\captionsetup{justification=centering}
\caption {Domain shifts used in our experiments}
\scalebox{1}{
\begin{tabular}{p{.7cm}p{6.6cm}}
\hline
 & \textbf{Domain shift} \\ \hline
\textbf{I} & Observation viewpoint, object properties, morphology of the manipulator  and scene background remain the same in the demonstration and learning domain \\

\textbf{V} & Observation viewpoint is different between the demonstration and the learning domain; Other factors remain unchanged \\ 
\textbf{Obj} & Objects with different colour (for pushing, reaching and hammering tasks) or shape (hammering task) used in the learning domain \\ 
\textbf{Obj+V} & Both the viewpoint of observation and object properties vary between the demonstration and the learning domains \\ 
\textbf{BG} & Background clutter is introduced to the scene in learning domain, which was not present during the demonstration \\ 
\textbf{M} & Manipulators with different morphologies used in the demonstration and the learning domain. Demonstrations with a human hand (reaching and pushing tasks) and with a manipulator with a different morphology (hammering task) used. \\ \hline
\end{tabular}
}
\label{table:experiment-setups}
\end{table}

\begin{table}[!t]
\tablestyle[default]
\rmfamily
\centering
\captionsetup
{justification=centering}
\caption {Task definitions and completion measures}
\scalebox{1}{
\begin{tabular}{|p{1.7cm}|p{1.9cm}|p{3.3cm}|}
\hline
\textbf{Task} & \textbf{Description} & 
\textbf{Task completion measure} 
\\ \hline
Reaching  (Simulation) & Reach a target zone & 1-(final distance / initial distance between the center of the manipulator and the center of the target zone) 
\\ 
\hline
Pushing (Simulation \& real robot) & Push an object into the target zone & 1-(final distance/initial distance between the centers of the target zone and the pushed object) \\ \hline
Hammering (Real robot) & Hammer the target object & 1-(minimum distance / initial distance between the hammer and the object during the execution) \\ 
\hline
Sweeping (Real robot)  & Sweep crumpled cardboard pieces to the dustbin &  The cardboard pieces in the dustbin before execution / the cardboard pieces in the dustbin before execution) \\ 
\hline
Striking (Real robot)  & Strike down a block of cubes & 1-(minimum distance / initial distance between the blocks and manipulator during execution) \\ 
\hline
\end{tabular}
}
\label{table:tasks}
\end{table}

To explore the resilience of our method to shifts between the demonstrator and learner domains, we conducted the experiments with six different domain shifts, as defined in Table~\ref{table:experiment-setups}. The tasks used are reaching and pushing in simulation and pushing, hammering, sweeping and striking for the real robot experiment. The task definitions and completion measures are given in Table \ref{table:tasks}. Note that the task completion measures are only used for evaluating the performance of O\textsubscript{2}A and not used during learning.

\subsection{Simulation experiment}

We set up the simulation learning domain with a 3DOF robotic manipulator for reaching and pushing using OpenAI Gym~\cite{brockman2016openai} and the MuJuCo physics engine\cite{mujoco}. In each setup (characterising a domain shift), we collect a single demonstration in the real world and run DDPG algorithm 10 times. Each run has 20 episodes per run and the number of steps per episode are 60 and 160 for reaching and pushing respectively. The network architectures and hyperparameters used by DDPG are given in the supplementary material Section D. For each run, the DDPG returns a control policy that corresponds to the maximum reward obtained.  After training, we pick the top-2~\cite{Henderson2018DeepRL} control policies with the highest rewards, and the task completion measures are calculated. The top control policies were selected to avoid policies from poorly performing runs affecting the overall  performance. The output of the control policy are the robotic controls with a size of three corresponding to each of the joints. The robotic controls could be torques, joint positions or velocities of the manipulator. In our experiment we have used joint positions. We perform the experiment with action vectors extracted from both pool5 and fc6 layers of NN:UCF101 model. Figure~\ref{fig:demons_n_learned} shows snapshots of the demonstration and execution of the corresponding learned policy for selected setups. Videos of the  simulation experiment results, including demonstrations are available in the project website.

\begin{figure*}[!t]
   \centering
   \includegraphics[scale=.38]{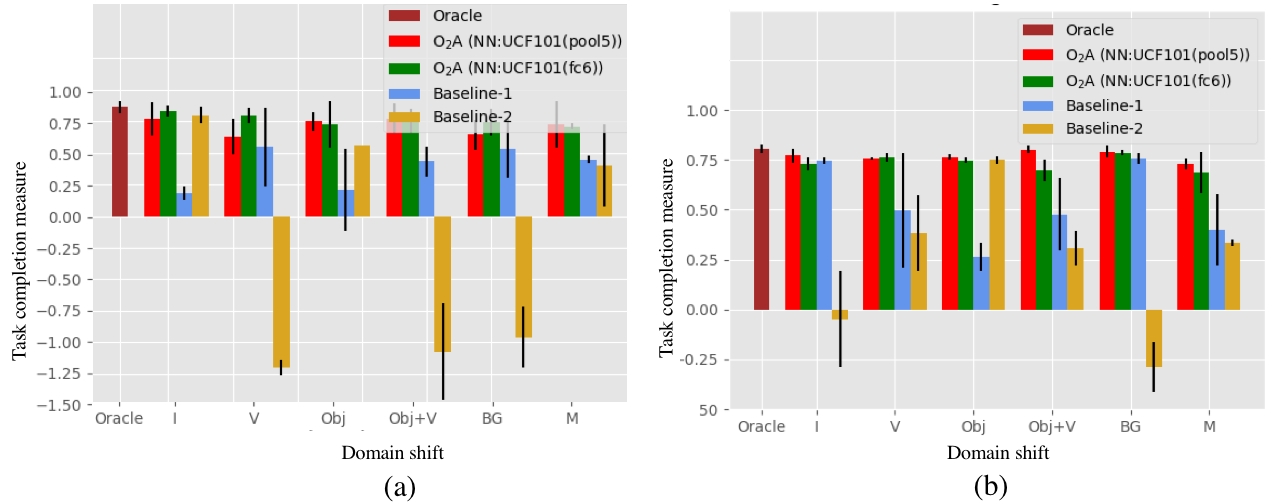}
    \caption{ Task completion measures for the task of (a) reaching and (b) pushing in the simulation experiment. O\textsubscript{2}A outperforms both the baselines and has performance comparable to the Oracle under all domain shifts. The Oracle score is shown only once since it is unaffected by the domain shifts (refer to Table \ref{table:experiment-setups} for domain shift definitions). }
    \label{fig:sim_res_combined}
\end{figure*}

We compare our method with an oracle and two baseline approaches. The oracle is trained by using the corresponding task completion measure specified in Table~\ref{table:tasks} as the reward, in place of a reward derived from action vectors. It represents the upper bound on performance. The two baselines represent a video clip by averaging a \lq static\rq~ representation for each frame, in contrast to the spatio-temporal representation used in O\textsubscript{2}A. Rewards are then generated using these representations. In Baseline-1, features from the output of the last convolutional layer of the ImageNet~\cite{simonyan2014very} pre-trained VGG-16 network are used and in Baseline-2, HOG~\cite{dalal2005histograms} features are used. The average of the task completion measures for the top two control policies for oracle, O\textsubscript{2}A and the baseline approaches  are plotted in Figure \ref{fig:sim_res_combined}. The learned policies from O\textsubscript{2}A were successful in performing the demonstrated task under different domain shifts with good task completion measures. It also significantly outperforms both baseline approaches and has a comparable performance to the oracle. 

Additionally we also analysed the quality of the  rewards generated in O\textsubscript{2}A, the baseline approaches and the oracle using correlation of rewards. Results are reported in the supplementary material Section D.

\subsubsection{Trajectory Maps}

\begin{figure}[!h]
   \centering
   \smallskip
   \medskip
   \includegraphics[scale=.35]{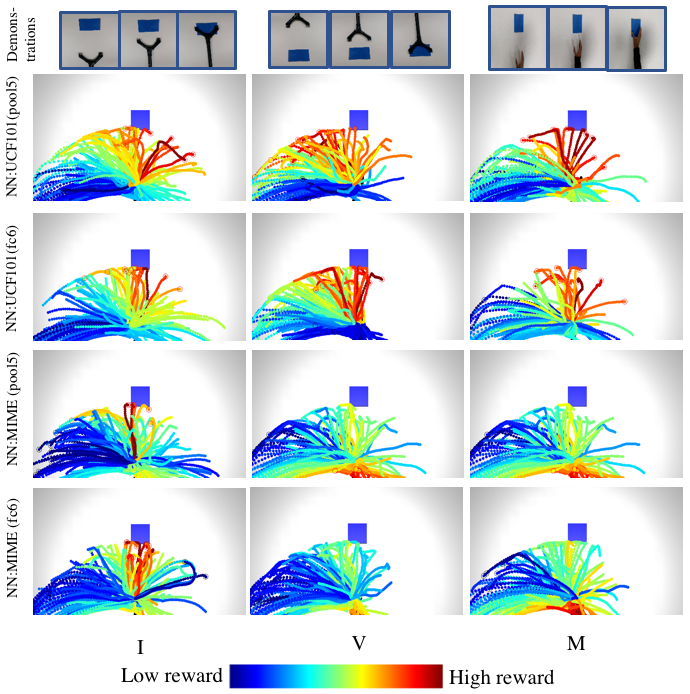}  
    \caption{
    Trajectory maps obtained during reinforcement learning of the task of reaching, when O\textsubscript{2}A action vector extractors are pre-trained with UCF101 dataset (NN:UCF101 (pool5, fc6)) and with MIME dataset (NN:MIME (pool5, fc6)). NN:UCF101 provides high rewards for desired trajectories for all the domain shifts (I, V, M). However, NN:MIME performs poorly when viewpoint of observation (V) and morphology of the manipulator (M) changes.}
    \label{fig:traj_map_all}
\end{figure}

Here we plot the trajectories followed by the robotic manipulator in each episode during reinforcement learning of the task. This visualisation will help to understand, if high rewards are obtained for desired trajectories while learning the demonstrated task. The trajectories are coloured with a colour scale corresponding to normalised reward values obtained during task learning.  

We also show the results when O\textsubscript{2}A action vector extractors are pre-trained with a manipulation task dataset, the Multiple Interactions Made Easy (MIME) dataset \cite{Sharma2018}. MIME dataset consists of 8260 videos of 20 commonly seen robotic manipulation tasks, executed by a human as well a Baxter robot. This model is referred to as \lq NN:MIME\rq. The aim is to study how well O\textsubscript{2}A perform when pre-trained on a task specific manipulation dataset compared to a generic dataset. Relevant results are shown in Figure \ref{fig:traj_map_all} and the rest can be found in the supplementary material Section D.

The results indicate that, when NN:UCF101 is used, high rewards are generated for desired trajectories for all domain shifts. However NN:MIME performs poorly for changes in viewpoint and manipulator used. An insight into this is that, even though the MIME dataset consist of large number of manipulation task examples, the variations in terms of viewpoints and manipulators used are limited. In contrast UCF101 contains examples with extensive range of variations in domain settings like viewpoint and manipulator morphology.

\begin{figure*}[!t]
   \centering
   \includegraphics[scale=.38]{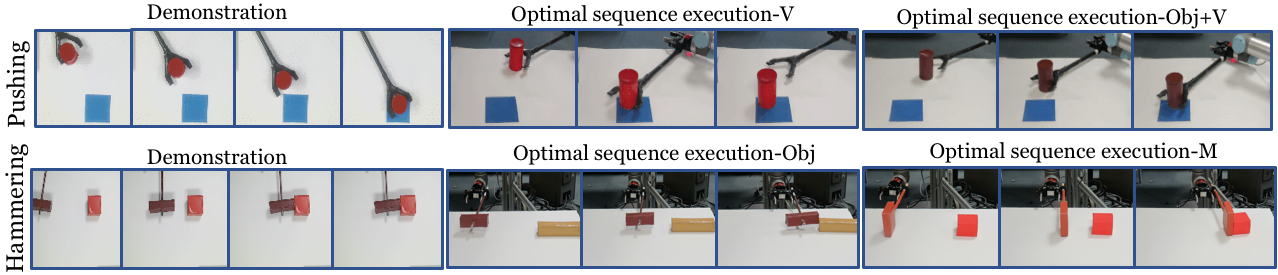}
    \caption{Snapshot of the demonstration and execution of the corresponding optimal control sequences obtained for selected domain shifts from the real robot experiment (Results shown for action vectors extracted from pool5 layer of NN:UCF101 model).}
    \label{fig:real_res}
\end{figure*}

\begin{figure}[!h]
   \centering
   \includegraphics[scale=.25]{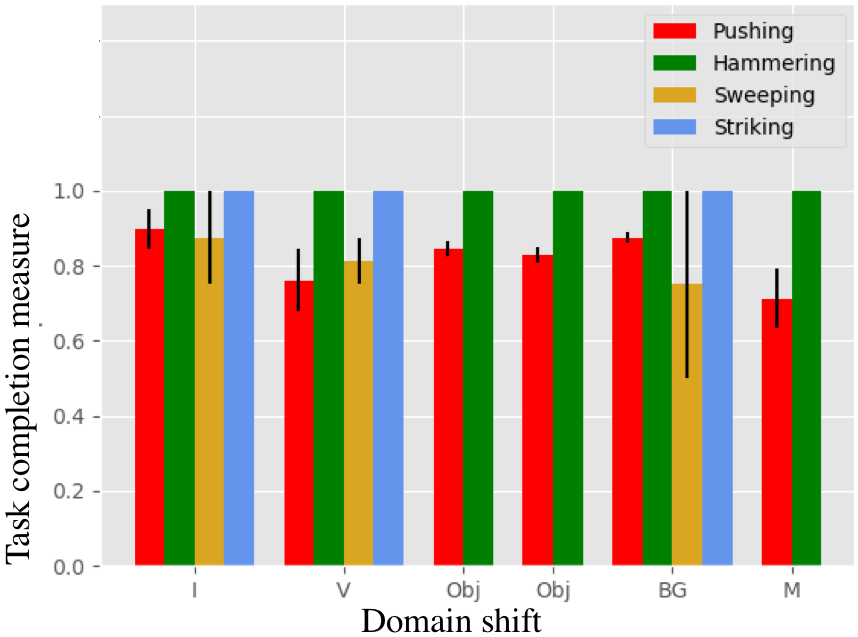}
    \caption{Task completion measures for the task of pushing, hammering, sweeping and striking in the real robot experiment. The result shows that O\textsubscript{2}A performs well under different domain shifts on a real robot.}
    \label{fig:real_e2345}
\end{figure}

\subsection{Real robot experiment}
\label{sec:exp_results_real_robot}

For the real robot experiment, we use a 6-DOF  UR5 robotic arm with different end-effectors suitable for each task. All 6 domain shifts (see Table \ref{table:experiment-setups}) are used for the pushing and hammering tasks. Whereas, only three domain shifts (I,V and M) are used for the sweeping and striking tasks, since others did not have meaning for these tasks. We only used features from pool5 layer of NN:UCF101 model as the action vector, due to the high cost of running the real robot experiment. Implementation details of the STO algorithm used to generate the optimal sequence of controls are given in the supplementary material Section E.

Each experiment is run two times with 10 iterations each. In Figure \ref{fig:real_res} the snapshots of executions of optimal control sequences obtained for the selected setups are given. The average task completion measures for the optimal control sequences are shown in Figure \ref{fig:real_e2345}. Our method achieves good task completion measures for different domain shifts. This shows the effectiveness of O\textsubscript{2}A in learning tasks on a real robot. Videos of all the results of the real robot experiments, including demonstrations are available in our project website.

\section{Conclusion} \label{conclusion}
We have presented O\textsubscript{2}A, a method for observational learning of robotic manipulation tasks from a single (one-shot) demonstration. The method works by extracting a perceptual representation (the action vector) from videos using pre-trained action vector extractor. Our analysis shows
that the pre-trained action vector extractor can generalise to unseen robotic manipulation tasks. Also experiments in simulation and with a real robot show that O\textsubscript{2}A can perform well under different domains shifts and outperforms baseline approaches.

A limitation in our work is the number of trial executions required to learn a task. It would be interesting to see if we can map the action vector from the demonstration directly to a initial near optimal solution. Another future direction will be to use additional sensing modalities like touch or audio for situations where the demonstrations are not visually observable (e.g. due to occlusion). Also, it would be interesting to study pre-training on generic action datasets for other robotic manipulation problems. Such pre-training could potentially address the lack of large ImageNet~\cite{simonyan2014very} like datasets of robotic manipulation task videos. Finally it would be exciting to extend O\textsubscript{2}A to multi-step manipulation tasks. One approach to tackle this could be to decompose these tasks into single-step tasks learnt using the current method, within a curriculum learning framework.

\textbf{Acknowledgement}: The authors would like to appreciate Mohammed abdellathif, Matteo Leonetti, Mehmet Doger, Wissam Bejjani, Rebecca Stone, Mohammed Alghamdi, Hanh Tran and Mohammad Kaykanloo for fruitful discussions on our work. 


\bibliographystyle{ieeetr}

\clearpage
\appendix
\onecolumn

\begin{center}
\end{center}
\medskip
\medskip
\medskip
\medskip
\medskip

\begin{center}
\textbf{\large {O\textsubscript{2}A: One-shot Observational learning with Action vectors \\- Supplementary Material -}}
\end{center}

\medskip
\medskip
\medskip
\medskip
\medskip
\medskip
\medskip

\section{Literature review}
\label{app:A}

Observational learning methods in existing  literature are compared in Table \ref{table:literature_obs_learn}.
O\textsubscript{2}A requires only a single demonstration to learn new tasks. It does not use any robot data for training the action vector extractor. And also works well under different all domain shifts.

\begin{table}[!h]
\centering
\tablestyle[default]
\rmfamily
\captionsetup{justification=centering}
\caption{Comparing observational learning methods in existing  literature. }
\label{table:literature_obs_learn}
\scalebox{.8}{
\begin{tabular}{c p{7.2cm} c c c c c}
\hline
Reference & \begin{tabular}[c]{@{}c@{}}No: of video demonstrations required per task \\ (Including to train  the feature extractor/s) \end{tabular}                                                                                                         & \begin{tabular}[c]{@{}c@{}}Is robot data \\ required  for training \\ the feature  extractor/s ?\end{tabular} & \begin{tabular}[c]{@{}c@{}}Viewpoint \\ invariant ?\end{tabular} & \begin{tabular}[c]{@{}c@{}}Invariant to \\ changes of \\ object \\ properties ?\end{tabular} & \begin{tabular}[c]{@{}c@{}}Invariant to \\ changes in \\ scene\\  background ?\end{tabular} & \begin{tabular}[c]{@{}c@{}}Invariant to \\ changes of  \\ morphology of \\ the manipulator ?\end{tabular} \\ \hline
      \cite{sermanet2018time}    & $\sim$40 min of human demonstrations  +    $\sim$20 min random robot manipulation data                                                    & \checkmark                                                                                                                                                                    & \checkmark                                                            & \checkmark                                                                                     & \checkmark                                                                                    & \checkmark                                                                                                     \\ 
     \cite{stadie2017}    & An expert policy is used instead of direct demonstrations                                                                                                                                                                                   & \checkmark                                                                            & \checkmark                                                            & \checkmark                                                                                     & \checkmark                                                                                    & \checkmark                                                                                                     \\ 
      \cite{Liu2018}    & $\sim$60-3000 human demonstrations using additional tools                                                                                                                                                                    & \ding{55}                                                                             & \checkmark                                                            & \checkmark                                                                                     & \checkmark                                                                                    & \ding{55}                                                                                                      \\ 
     \cite{Smith2019}     & $\sim$20-30 human demonstrations + $\sim$300-500 random human and robot images                                                                                                   & \checkmark                                                                            & \ding{55}                                                             & NA                                                                                    & NA                                                                                   & \checkmark                                                                                                     \\ 
      \cite{Sharma2019}    & $\sim$230 human demonstrations + corresponding robotic joint angle data                                                                                                                                                                       & \checkmark                                                                            & \checkmark                                                            & \checkmark                                                                                     & NA                                                                                   & \checkmark                                                                                                   \\ 
     \cite{Sharma2018}     & $\sim$200-400 human demonstrations + corresponding robotic joint angle data                                                                                                                                                                   & \checkmark                                                                            & \checkmark                                                            & \checkmark                                                                                     & NA                                                                                   & \checkmark                                                                                                     \\ 
       \cite{sermanet2016unsupervised}     &   $\sim$12 human demonstrations                                                                                                                                                                           & \ding{55}                                                                             & \ding{55}                                                             & \checkmark                                                                                     & NA                                                                                   & \checkmark                                                                                                     \\ 
       \cite{Jia2020}   & $\sim$50-100 human demonstrations                                                                                                                                                                         & \ding{55}                                                                             & \checkmark                                                            & \checkmark                                                                                     & \checkmark                                                                                    & \checkmark                                                                                                     \\ 
      \cite{Tow2017}    & Uses both human and robot task demonstrations (exact numbers unknown)                                                                                                                                                                                           & \checkmark                                                                            & \checkmark                                                            & NA                                                                                    & NA                                                                                   & \checkmark                                                                                                     \\ 
     \cite{Yang2019}     & $\sim$2990 human demonstrations                                                                                                         & \ding{55}                                                                             & \checkmark                                                            & \checkmark                                                                                     & \checkmark                                                                                    & \checkmark                                                                                                     \\ 
    \cite{yu2018one}      & 1 (But uses closely related supplementary task demonstrations. Requires $\sim$600-1200 robot and $\sim$600-1200 human demonstrations per task) & \checkmark                                                                            & \checkmark                                                            & \checkmark                                                                                     & \checkmark                                                                                    & \checkmark                                                                                                     \\ 
     \cite{yu2018two}     & 1 (But requires large number of action primitive demonstrations. $\sim$600-1200 robot and $\sim$600-1200 human demonstrations per action primitive) & \checkmark                                                                            & NA                                                             & \checkmark                                                                                     & \checkmark                                                                                    & \checkmark                                                                                                     \\ \hline
    O\textsubscript{2}A      & Only 1 demonstration   (human demonstration with or without using additional tools)                                                                                                                                                                                      & \ding{55}                                                                             & \checkmark                                                            & \checkmark                                                                                     & \checkmark                                                                                    & \checkmark                                                                                                     \\ \hline
\end{tabular}
}
\end{table}
\clearpage
\twocolumn

\section{Action vector extractor: Model architecture and pre-training details}
\label{app:B}

The architecture of the 3D-CNN model used is given in Table \ref{Table:c3d}. Variables NC and BS denotes, the number of classes and batch size respectively. The details of pre-training the model for action classification are given in Table \ref{table:actiontraining}.

\begin{table}[!h]
\tablestyle[default]
\rmfamily
\centering
\captionsetup{justification=centering}
\caption{3D-CNN model used for action vector extraction used in our experiments}
\scalebox{.67}{
\label{Table:c3d}
\begin{tabular}{|l|l|l|l|l|}
\hline
\textbf{Layer} & \textbf{Type}          & \textbf{Kernel size}       & \textbf{Input size}                     & \textbf{Output size }           \\ \hline
\multicolumn{5}{|l|}{} \\ \hline
conv1      & Conv3D        & (3, 3, 3)          & (BS, 16, 112, 112, 3)        & (BS, 16, 112, 112, 64) \\ \hline
pool1      & MaxPooling3D  & (1, 2, 2)              & (BS, 16, 112, 112, 64) & (BS, 16, 56, 56, 64)   \\ \hline
\multicolumn{5}{|l|}{} \\ \hline
conv2      & Conv3D        & (3, 3, 3)         & (BS, 16, 56, 56, 64)         & (BS, 16, 56, 56, 128)  \\ \hline
pool2      & MaxPooling3D  & (2, 2, 2)              & (BS, 16, 56, 56, 128)        & (BS, 8, 28, 28, 128)   \\ \hline
\multicolumn{5}{|l|}{} \\ \hline
conv3a     & Conv3D        & (3, 3, 3)         & (BS, 8, 28, 28, 128)         & (BS, 8, 28, 28, 256)   \\ \hline
conv3b     & Conv3D        & (3, 3, 3)         & (BS, 8, 28, 28, 256)         & (BS, 8, 28, 28, 256    \\ \hline
pool3      & MaxPooling3D  & (2, 2, 2)              & (BS, 8, 28, 28, 256)         & (BS, 4, 14, 14, 256)   \\ \hline
\multicolumn{5}{|l|}{} \\ \hline
conv4a     & Conv3D        & (3, 3, 3)         & (BS, 4, 14, 14, 256)         & (BS, 4, 14, 14, 512)   \\ \hline
conv4b     & Conv3D        & (3, 3, 3)         & (BS, 4, 14, 14, 512)         & (BS, 4, 14, 14, 512)   \\ \hline
pool4      & MaxPooling3D  & (2, 2, 2)              & (BS, 4, 14, 14, 512)         & (BS, 2, 7, 7, 512)     \\ \hline
\multicolumn{5}{|l|}{} \\ \hline
conv5a     & Conv3D        & (3, 3, 3)         & (BS, 2, 7, 7, 512)           & (BS, 2, 7, 7, 512)     \\ \hline
conv5b     & Conv3D        & (3, 3, 3)         & (BS, 2, 7, 7, 512)           & (BS, 2, 7, 7, 512)     \\ \hline
zeropad5   & ZeroPadding3D & \multicolumn{1}{c|}{-} & (BS, 2, 7, 7, 512)           & (BS, 2, 8, 8, 512)     \\ \hline
pool5      & MaxPooling3D  & (2, 2, 2)              & (BS, 2, 8, 8, 512)           & (BS, 1, 4, 4, 512)     \\ \hline
flatten1  & Flatten       & \multicolumn{1}{c|}{-} & (BS, 1, 4, 4, 512)           & (BS, 8192)             \\ \hline
\multicolumn{5}{|l|}{} \\ \hline
fc6        & Dense         & (4096, 8192) & (BS, 8192)                   & (BS, 4096)             \\ \hline
fc7        & Dense         & (4096, 4096) & (BS, 4096)                   & (BS, 4096)             \\ \hline
fc8        & Dense         & (NC, 4096) & (BS, 4096)                   & (BS,  NC)  \\ \hline
\end{tabular}
}
\end{table}

\begin{table}[!h]
\tablestyle[default]
\rmfamily
\centering
\captionsetup{justification=centering}
\caption{Pre-training details for the NN:UCF101 model}
\label{table:actiontraining}
\scalebox{.96}
{
\begin{tabular}{p{5cm}cc}
\hline
\textbf{Number of classes (NC)  }                                                        & 101                                                                     \\ 
\textbf{Batch size (BS)}                                                                 & 16                                                                      \\ 
\textbf{Input size }                                                                     & (16, 16, 112, 112, 3)        \\ 
\textbf{Output size}                                                                     & (16, 101)                                                       \\ 
\textbf{GPUs used }                                                                      & 2 x Nvidia K80                                              \\ 
\textbf{Training time }                                                                  & 48 hrs                                                           \\ 
\textbf{Optimizer }                                                                      & ADAGRAD \cite{duchi2011adaptive}                                   \\ 
\textbf{Learning rate}                                                                   & 0.001                                                                \\
\textbf{Number of training examples}                                                     & 9,990                                                                 \\ 
\textbf{Number of validating examples}                                                    & 3,330                                                                \\ 
\begin{tabular}[c]{@{}l@{}}\textbf{Total number of}\\  \textbf{trainable parameters}\end{tabular} & 78,409,573                                                     \\ 
\textbf{Number of epochs}                                                                & 119                                                               \\ 
\textbf{Best validation accuracy }                                                            & 60.72\%                                                              \\ \hline
\end{tabular}
}
\end{table}

\section{Clustering analysis}
\label{app:C}

We conduct the experiment to identify which one of the final layers of NN:UCF101 provides the best action vector for manipulation tasks. We use the quality of the clusters in the action vector space, as a measure to understand how task-aware and domain-invariant are the action vectors from different layers of NN:UCF101 model. The more the action vectors are domain-invariant and task-aware, the better the clustering of the action vectors from the same class will be. To analyse the quality of the clusters, we use a standard clustering evaluation measure, the ARI \cite{hubert1985comparing} score. The ARI score measures the extent to which the predicted clustering corresponds to the and ground truth clusters by counting pairs that are assigned in the same or different clusters. ARI values are bounded by $[-1,1]$, where $-1$ is the lowest score, 0 indicates random clustering and 1 shows that the predicted clustering corresponds to the ground truth perfectly.

For the experiment, we extract the action vector from the pool5, fc6, fc7 and fc8 layers of the NN:UCF101 model, for all the 17 videos in LMD. The Baseline-R is obtained using features from the pool5 layer of the same NN:UCF101 model but initialised with random weights. The features extracted from each layer are then clustered using the K-means clustering algorithm. The value of K=3 is used, corresponding to the number of task classes. After clustering, the predicted cluster labels are evaluated against ground truth labels and ARI scores are calculated. The results of the experiment are tabulated in Table.~\ref{table:ARI}.

\begin{table}[!h]
\centering
\tablestyle
\rmfamily
\caption{ARI scores. Results show that the features from layer pool5 and fc6 of the NN:UCF101 model are best suited to be used as action vectors.}
\label{table:ARI}
\scalebox{.96}
{
\begin{tabular}{p{4cm}c}
\hline
\textbf{Layer}  &   \textbf{ \hspace{.7cm} ARI score\hspace {.7cm} } \\ 
\hline
\textbf{Baseline-R (Random weights)} &     0.07     \\ 
\textbf{pool5}  &   \textbf{0.26}     \\ 
\textbf{fc6} &   \textbf{0.34}  \\ 
\textbf{fc7} &   0.19  \\ 
\textbf{fc8} &   0.14  \\
\hline
\end{tabular}
}
\end{table}

The ARI value for Baseline-R is close to zero as expected and gives us the baseline to compare with. The ARI score increases when features from pool5 to fc6 layers are used as the action vector, but drops for the final fc7 and fc8 layers. The results indicate that the features from pool5 and fc6 layers of the NN:UCF101 model are the most suitable to be used as the action vector.

\section{Simulation experiment}

\subsection{DDPG algorithm}
\label{app:D}

Details of the DDPG algorithm used in the simulation experiment are given here. We use architectures identical to \cite{Lillicrap2015ContinuousCW} for the actor and critic networks. The hyper-parameters used are given in Table \ref{tab:rl_train_traj}. 

\begin{table}[h]
\centering
\tablestyle[default]
\rmfamily
\caption{DDPG hyperparameters used}
\label{tab:rl_train_traj}
\scalebox{.96}
{
\begin{tabular}{p{6cm}c}
\hline
\multicolumn{1}{l}{\textbf{ Hyperparameter}} & \multicolumn{1}{l}{\textbf{Value}} \\ \hline
Actor learning rate                   & 0.0001                     \\
Critic learning rate                  & 0.001                      \\
State size (S\_size)                   & 4608                       \\
Action size (A\_size)                  & 3                          \\
Optimiser used                        & ADAM                       \\
Gamma ($\gamma$)                      & 0.99                       \\
Tau ($\tau$)                          & 0.001                      \\
Mini batch size (MB\_size)             & 64                         \\
Replay buffer size                    & 10,000   \\ \hline                 
\end{tabular}
}
\end{table}

\subsection{Correlation of rewards}

\begin{table*}[!b]
\centering
\tablestyle[default]
\rmfamily
\smallskip
\medskip
\caption {\centering Pearson correlation coefficients between the rewards from the Oracle, and from O2A and two baselines. The coefficients are generally highest and positive for O\textsubscript{2}A rewards compared to baseline approaches.}
\scalebox{.98}{
\begin{tabular}{|c|c|c|c|c|c|c|}
\hline
\multicolumn{7}{|c|}{\textbf{Task 1: Reaching}}                                                                       \\ \hline
           & I          & V          & Obj        & Obj+V        & BG          & M          \\ \hline
O\textsubscript{2}A (NN:UCF101 (pool5)) & \textbf{.8567}\textpm\textbf{.0079}  & {.7807}\textpm{.0531}  & \textbf{.8209}\textpm\textbf{.0157}   & {.6448}\textpm{.2146}  & {.7736}\textpm{.0007}   & \textbf{.9605}\textpm\textbf{.0048}           \\ \hline
O\textsubscript{2}A (NN:UCF101 (fc6)) & .8318\textpm.0600 & \textbf{.7911}\textpm\textbf{.0588} &.8199\textpm.0718
 & \textbf{.8620}\textpm\textbf{.0713}
 & \textbf{.8108}\textpm\textbf{.1126}
 & .8761\textpm.0032
\\ \hline
Baseline-1 & .5872\textpm.1744  & .4069\textpm2361   & .6112\textpm.2612   & .6099\textpm0901  & .5289\textpm.0189  &   .0487\textpm.0448      \\ \hline
Baseline-2 & .7387\textpm.0681  & -.8106\textpm.0086 & .7115\textpm.1272   & -.8189\textpm.0501 & -.5738\textpm.0337 &   .1256\textpm.0629         \\ \hline
\multicolumn{7}{|c|}{\textbf{Task 2: Pushing}}                                                                        \\ \hline
O\textsubscript{2}A (NN:UCF101 (pool5))& .9345\textpm.0034    & \textbf{.9413}\textpm\textbf{.0362}  & \textbf{.6943}\textpm\textbf{.1419}     & \textbf{.8650}\textpm\textbf{.0847}   & .8552\textpm.0677  & {.6594}\textpm{.1834} \\ \hline
O\textsubscript{2}A (NN:UCF101 (fc6))& .8037\textpm.1125
 &.8826\textpm.0239 & .6898\textpm.1927& .8179\textpm.0702
& \textbf{.9147}\textpm\textbf{.0099} &
\textbf{.8489}\textpm\textbf{.0987} \\ \hline
Baseline-1 & \textbf{.9372}\textpm\textbf{.0270}  & .8908\textpm.0615  & .5817\textpm.3124 & .7488\textpm.0631  & {.8978}\textpm{.0704}   & .5797\textpm.1141  \\ \hline
Baseline-2 & .0173\textpm.4550 & -.1346\textpm.3410  & .5900\textpm.1625   & -.4352\textpm.1292 & -.5386\textpm.1243 & .3700\textpm.5195 \\ \hline
\end{tabular}
}
\label{table:coefficients}
\end{table*}

We further analysed the quality of the  rewards generated in O\textsubscript{2}A, the baseline approaches and the Oracle. To compare, we calculate the Pearson correlation coefficient \cite{benesty2009pearson} between the episodic perceptual rewards ( O\textsubscript{2}A, baselines) and the Oracle rewards for the top two runs. A high positive correlation (typically $>$ 0.5 \cite{tipping1999probabilistic}) indicates that the perceptual rewards are as good as the Oracle rewards. All the results are tabulated in Table ~\ref{table:coefficients}. From the results, the correlation coefficients are greater than 0.5 in all the cases for O\textsubscript{2}A, indicating that our rewards are as accurate as the Oracle rewards. Also, the correlation is higher and positive compared to the baselines for a range of domain shifts showing the superior performance of our method. 

\subsection{Trajectory maps}
\begin{figure}[!h]
   \centering
   \smallskip
   \medskip
   \includegraphics[scale=.35]{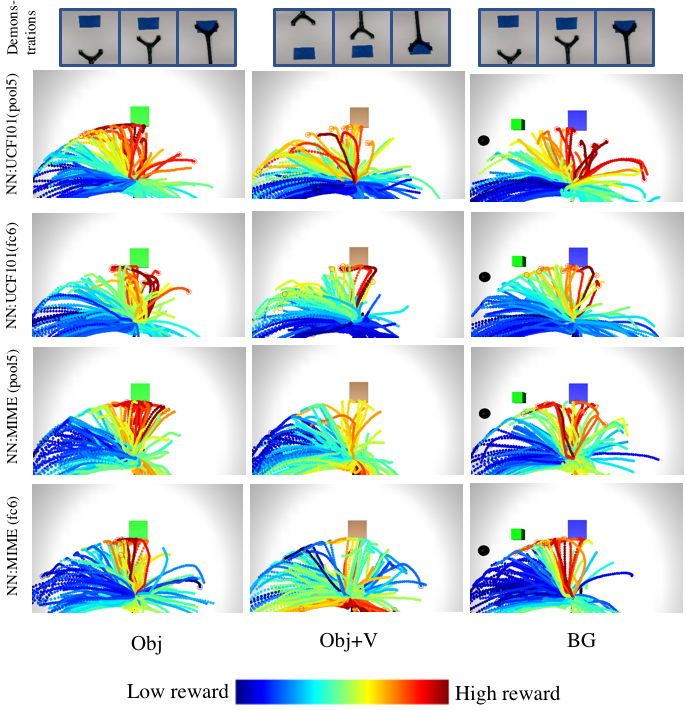}  
    \caption{Trajectory maps while learning the task of reaching, when O\textsubscript{2}A action vector extractors are pre-trained with UCF101 dataset (NN:UCF101 (pool5, fc6)) and with MIME dataset (NN:MIME (pool5, fc6)). O\textsubscript{2}A pre-trained with UCF101 provides high rewards for desired trajectories for all the domain shifts (Obj, Obj+V, BG). However, O\textsubscript{2}A with MIME dataset fails when viewpoint of observation (Obj+V) changes.}
    \label{fig:traj_map_all_suppli}
\end{figure}

The trajectory maps for the rest of the setups (Obj,Obj+V and BG) for the task of reaching are give in Figure \ref{fig:traj_map_all_suppli}.

\section{STO algorithm implementation}
\label{app:E}

Briefly,  we  begin  with  an  initial  candidate  control  sequence.  We  execute  this  sequence  using the  manipulator  to  generate  an  initial  cost.  Thereafter,  at each  iteration  we  create  8  random  control  sequences  by adding  Gaussian  noise  to  the  candidate  sequence  from  the previous  iteration  and  execute  them  using  the  real  robot. At  the  end  of  each  iteration,  we  pick  the  control  sequence with  the  minimum  cost. Then  set  it  as  the  new  candidate sequence thereby  iteratively  reducing  the  cost. The initial control sequence is  initialised by providing a near solution path, following common practices in literature \cite{sermanet2016unsupervised}. However, a more sophisticated algorithm can be used to obtain the optimal control sequence without this.

\end{document}